# Hypergraph convolutional neural network-based clustering technique

**Loc H. Tran[1,2], Nguyen Trinh[1,2], Linh H. Tran[1,2]**
[1]Department of Electronics, Ho Chi Minh City University of Technology (HCMUT), Ho Chi Minh City, Vietnam
[2]Department of Electronics, Vietnam National University Ho Chi Minh City (VNU-HCM), Ho Chi Minh City, Vietnam

**ABSTRACT**

This paper constitutes the novel hypergraph convolutional neural network-based clustering technique. This technique is employed to solve the clustering problem for the Citeseer dataset and the Cora dataset. Each dataset contains the feature matrix and the incidence matrix of the hypergraph (i.e., constructed from the feature matrix). This novel clustering method utilizes both matrices. Initially, the hypergraph auto-encoders are employed to transform both the incidence matrix and the feature matrix from high dimensional space to low dimensional space. In the end, we apply the k-means clustering technique to the transformed matrix. The hypergraph convolutional neural network (CNN)-based clustering technique presented a better result on performance during experiments than those of the other classical clustering techniques.

*Corresponding Author:*

Linh H. Tran
Department of Electronics, Ho Chi Minh City University of Technology
Ho Chi Minh City, Vietnam
Email: linhtran@hcmut.edu.vn

## 1. INTRODUCTION

A crucial problem in deep learning and machine learning research areas is clustering. It is a technique that separates data points/samples into groups/clusters so that these data points/samples are more related to other data points/samples in the same groups/clusters than those in the other groups/clusters. Its applications are huge such as mobility pattern clustering [1], text clustering [2]–[5], and customer segmentation [6]–[9].

It is worth to mention about the motivation of this one specific clustering problem, which is the text clustering problem of our company. In detail, we would like to partition Facebook users into their appropriate groups/clusters (i.e., depending on the contents that they are talking about). For example, there are groups of Facebook users talking about games, there are groups of Facebook users talking about music, and there are groups of Facebook users talking about religions.

The applications of this text clustering problem are huge such as: i) Our influencers/streamers can create the "appropriate" content for their fans/users; ii) This text clustering problem leads to the implementation of the recommendation/matching systems (for example, fans/influencers, and fans/contents). We can employ the bi-partite graph to implement these recommendation/matching systems (i.e., bi-partite graph matching problem); and iii) Fake fans detection or anomaly/abnormal detection: for example, the fans/users belong to the group game, but they do not comment/feedback appropriately (i.e., they always talk about music).

Various clustering techniques are available from Python sklearn package [10] such as k-means [1], hierarchical clustering technique, and affinity propagation. These techniques could be used to handle many clustering problems. For notation, however, they can only be applied to feature datasets. Regardless, in this paper, we only employ k-means clustering technique as the "vanilla" or baseline technique because of three main reasons: i) It's very simple to implement, ii) It scales to large datasets, and iii) It guarantees to converge to final solutions.

Moreover, there are other clustering techniques (i.e. belong to different class of clustering techniques) that can also be employed to solve the clustering problem but can only be applied to network datasets such as spectral clustering [1], [11]–[13], and maximum modularity approach [14]–[20]. The weakness of these two classes of clustering techniques is obviously that they can only be applied to one type of dataset, which lead to information loss. About the case that we have both types of datasets such as the feature dataset and the network dataset, to assume that the samples in both datasets are the same, how can we apply the clustering technique to both datasets?

In this study, we develop a novel clustering method. This method utilizes both the feature dataset and the network dataset. We discuss the details below. Initially, we employ the graph auto-encoders proposed by Thomas Kipf [21]–[25] to transform both the network dataset and the feature dataset from high dimensional space to low dimensional space. In the end, the k-means clustering technique is applied to the transformed dataset. This novel clustering technique is called the graph convolutional neural network-based clustering technique. There are some main advantages of this novel clustering technique such as:

- The memory/space required to store the data is lowered. This leads to the low space complexity.
- The low space complexity leads to less computational and training time of the classification/clustering techniques (i.e. the low time complexity).
- The noises and redundant features are removed from the feature dataset and the network dataset. This leads to the high performance of the clustering technique. The claim is backed in section 4.
- Both network dataset and feature dataset are utilized. This leads to no information loss.

However, there is one weakness associated with the graph convolutional neural network (CNN)-based clustering technique. In other words, assuming the pairwise relationships among the objects/entities/samples in this graph representation is not complete. Let's consider the case that we would like to partition/segment a set of articles into different topics [26], [27]. Initially, we employ the graph data structure to represent this dataset. The vertices of the graph are the articles. Two articles are connected by an edge (i.e., the pairwise relationship) if there is at least one author in common. Finally, we can apply clustering technique to this graph to partition/segment the vertices into groups/clusters.

Obviously, we ignore the information whether one specific author is the author of three or more articles (i.e., the co-occurrence relationship or the high order relationship) in this graph data structure. This leads to information loss and eventually, poor performance (i.e., the low accuracy) of the clustering technique. To overcome this difficulty, we try to employ the hypergraph data structure to represent the above relational dataset. In detail, in this hypergraph data structure, the articles are the vertices, and the authors are the hyper-edges. This hyper-edge could connect more than two vertices (i.e., articles). Please note that if only the feature dataset is given, we need to construct the hypergraph from the feature dataset. The discussion on how to build the hypergraph from the feature vectors is presented in section 3.

In detail, initially, we develop the hypergraph auto-encoders to transform both the hypergraph dataset (i.e., constructed from the feature dataset) and the feature dataset from high dimensional space to low dimensional space. In the end, the k-means clustering technique is applied to the transformed dataset. This novel clustering technique is called the hypergraph CNN-based clustering technique. For notation, these clustering techniques are un-supervised learning techniques. Hence, in this paper, we do not need labeled datasets.

The rest of the paper is organized as: section 2 defines the issue and presents the novel graph CNN-based clustering technique. Section 3 presents the novel hypergraph CNN-based clustering technique. Section 4 describes the Citeseer and the Cora dataset [28]. Then, section 4 compares the performance the hypergraph CNN-based clustering technique and the performances of the graph CNN-based clustering technique, the k-means clustering technique, and the spectral clustering technique tested on these two Citeseer and Cora datasets. Section 5 is the conclusion.

## 2. GRAPH CONVOLUTIONAL NEURAL NETWORK-BASED CLUSTERING TECHNIQUE
### 2.1. Problem formulation

Given a set of samples $\{x_1, x_2, \ldots, x_n\}$. $n$ is the total number of samples. The pre-defined number of clusters $k$. In detail, we have the adjacency matrix $A \in R^{n*n}$ that:

$$A_{ij} = \begin{cases} 1 \text{ if vertex } i \text{ is connected to vertex } j \\ 0 \text{ otherwise} \end{cases} \quad (1)$$

and the feature matrix $X \in R^{n*L_1}$ where $L_1$ is the dimensions of the feature vectors. Our objective is to output the clusters/groups $A_1, A_2, \ldots, A_k$ so that:

$$A_i = \{j | 1 \leq j \leq n \text{ and } j \text{ belongs to cluster } i\}. \tag{2}$$

### 2.2. Adjacency matrix is not provided

Suppose that we are given the feature matrix $X \in R^{n*L_1}$ but not the adjacency matrix $A \in R^{n*n}$. In this case, the similarity graph could be constructed from these feature vectors using the *k*-nearest neighbor (KNN) graph. To put it another way, sample *i* connects with sample *j* by an edge in a no direction: un-directed graph if sample *i* is among the KNN of sample *j* or sample *j* is among the KNN of sample *i*.

Section 4 describes in detail the method of building the similarity graph from feature vectors. In the end, we get the similarity graph. This graph is represented by the adjacency matrix *A*. Please note that this phase is required for spectral clustering techniques and graph CNN-based clustering techniques if only the feature matrix is provided.

$$A_{ij} = \begin{cases} 1 \text{ if sample } i \text{ is connected to sample } j \\ 0 \text{ if sample } i \text{ is not connected to sample } j \end{cases} \tag{3}$$

### 2.3. Graph convolutional neural network-based clustering technique

At-present, the set of feature vectors $\{x_1, x_2, \ldots, x_n\}$ and the relationships among the samples (represented by the adjacency matrix *A*) in the dataset are available. For notation, $x_i \in R^{1*L_1}, 1 \leq i \leq n$ and $A \in R^{n*n}$. We have $\hat{A} = A + I$, with *I* as the identity matrix. With $\hat{D}$ as the diagonal degree matrix of $\hat{A}$, we have $\hat{D}_{ii} = \sum_j \hat{A}_{ij}$. The result output (i.e., the embedding matrix) *Z* of the graph CNN could be defined as (4):

$$Z = \hat{D}^{-\frac{1}{2}} \hat{A} \hat{D}^{-\frac{1}{2}} ReLU\left(\hat{D}^{-\frac{1}{2}} \hat{A} \hat{D}^{-\frac{1}{2}} X \theta_1\right) \theta_2 \tag{4}$$

For notation, *X* is the input feature matrix and $X \in R^{n*L_1}$. $\theta_1 \in R^{L_1*L_2}$ and $\theta_2 \in R^{L_2*D}$ are two parameter matrices to be learned during the training process. For notation, *D* is the dimensions of the embedding matrix *Z*.

Next, we need to reconstruct the adjacency matrix *A* from *Z*. We obtain the reconstruction $A'$ representing the similarity graph using (5)

$$A' = sigmoid(ZZ^T) \tag{5}$$

The Rectified Linear Unit or the The *ReLU* operation is defined in (6):

$$ReLU(x) = \max(0, x) \tag{6}$$

A *sigmoid* function can be defined in (7):
$$sigmoid(x) = \frac{1}{1+e^{-x}} \tag{7}$$

For this graph auto-encoder model, we need to evaluate the cross-entropy error over all samples in the dataset:

$$L = -\frac{1}{n^2} \sum_{i=1}^{n} \sum_{j=1}^{n} A_{ij} \ln\left(sigmoid(z_i z_j^T)\right) + (1 - A_{ij})(1 - A_{ij} \ln\left(sigmoid(z_i z_j^T)\right) \tag{8}$$

Please note that $z_i$ is the row *i* vector of the embedding matrix *Z* and $z_i \in R^{1*D}$. We train the two parameter matrices $\theta_1 \in R^{L_1*L_2}$ and $\theta_2 \in R^{L_2*D}$ using the gradient descent method. Then, partition the samples $(z_i)_{i=1,\ldots,n}$ in $R^{1*D}$ with the k-means algorithm into *k* clusters/groups. In general, the graph CNN-based clustering technique can be shown in Figure 1.

### 2.4. Discussions about graph convolutional neural network-based clustering technique

From the section 2.3., unlike other clustering techniques such as k-means, we easily see that this proposed clustering technique (i.e., the graph CNN-based clustering technique) utilizes both the feature dataset and the network dataset. This is a very strong argument of this proposed technique. Since no information are lost, the performance of this novel clustering technique is expected to be higher than the performance of the other classic clustering techniques such as k-means. The claim is backed in section 4.

However, there is one major weakness of this proposed clustering technique. When new samples arrive, we cannot predict which clusters these samples belong to (unlike k-means clustering technique). In other words, we have to update the adjacency matrix, and we have to re-train our graph auto-encoder. Thus, this technique can only be considered as the offline clustering technique although its performance is a lot higher than the performance of the other online clustering techniques.

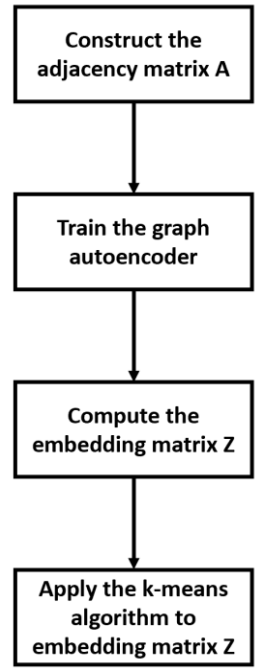

Figure 1. The graph convolutional neural network-based clustering technique

## 3. HYPERGRAPH CONVOLUTIONAL NEURAL NETWORK-BASED CLUSTERING TECHNIQUE
### 3.1. Problem formulation

Given a set of samples $\{x_1, x_2, \ldots, x_n\}$. $k$ is the pre-defined number of clusters. $n$ is the total number of samples. In details, we are given the incidence matrix $H \in R^{n*n}$ where:

$$H_{ij} = \begin{cases} 1 \text{ if vertex } i \text{ belongs to hyperedge } j \\ 0 \text{ otherwise} \end{cases} \quad (9)$$

and the feature matrix $X \in R^{n*L_1}$ where $L_1$ is the dimensions of the feature vectors. Our objective is to output the clusters/groups $A_1, A_2, \ldots, A_k$ where $A_i = \{j | 1 \leq j \leq n \text{ and } j \text{ belongs to cluster } i\}$.

### 3.2. Incidence matrix of the hypergraph is not provided

Suppose that we are given the feature matrix $X \in R^{n*L_1}$ but not the incidence matrix $H \in R^{n*n}$. In this case, using the KNN graph, we can create the incidence matrix $H$ from these feature vectors. Particularly, sample $i$ belongs to hyperedge $j$ if sample $i$ is among the KNN of sample $j$ or sample $j$ is among the KNN of sample $i$. We have $k$ set to be 5 in this paper. In the end, the incidence matrix $H$ which represents the hypergraph is obtained.

$$H_{ij} = \begin{cases} 1 \text{ if sample } i \text{ belongs to hyperedge } j \\ 0 \text{ if sample } i \text{ does not belong to hyperedge } j \end{cases} \quad (10)$$

For notation, this phase is required for hypergraph CNN-based clustering techniques if only the feature matrix is provided.

### 3.3. Hypergraph convolutional neural network-based clustering technique

We already have the incidence matrix $H$ of the hypergraph and the set of feature vectors $\{x_1, x_2, ..., x_n\}$. For notation, $x_i \in R^{1*L_1}, 1 \leq i \leq n$ and $H \in R^{n*n}$. With $w(e)$ as the weight of the hyper-edge $e$, and the $R^{n*n}$ diagonal matrix that contains the weights of all hyper-edges in its diagonal entries as $W$. From the weight matrix $W$ and the incidence matrix $H$, the degree of vertex $v$ and the degree of hyper-edge $e$ can be defined in:

$$d(v) = \sum_{e \in E} w(e) * h(v, e)$$

$$d(e) = \sum_{v \in V} h(v, e)$$

Let the two diagonal matrices that contains the degrees of vertices and the degrees of hyper-edges in their diagonal entries be $D_v$ and $D_e$, respectively. For additional notation, $D_v$ is the $R^{n*n}$ matrix and $D_e$ is the $R^{n*n}$ matrix. The final output (i.e., the embedding matrix) $Z$ of the hypergraph CNN is defined in (11):

$$Z = D_v^{-\frac{1}{2}} H W D_e^{-1} H^T D_v^{-\frac{1}{2}} ReLU \left( D_v^{-\frac{1}{2}} H W D_e^{-1} H^T D_v^{-\frac{1}{2}} X \theta_1 \right) \theta_2 \tag{11}$$

For notation, $X$ is the input feature matrix and $X \in R^{n*L_1}$. $\theta_1 \in R^{L_1*L_2}$ and $\theta_2 \in R^{L_2*D}$ are the two parameter matrices to be learned during the training process. For notation, the dimensions of the embedding matrix $Z$ is $D$.

Next, we need to reconstruct the incidence matrix $H$ from $Z$. We obtain the reconstruction $H'$ representing the hypergraph using (12)

$$H' = sigmoid(ZZ^T) \tag{12}$$

The Rectified Linear Unit and the $ReLU$ operation is defined in (13):

$$ReLU(x) = \max(0, x) \tag{13}$$

A $sigmoid$ function could be defined in (14):

$$sigmoid(x) = \frac{1}{1+e^{-x}} \tag{14}$$

In the case of this hypergraph auto-encoder model, we need to evaluate the cross-entropy error over all samples in the dataset:

$$L = -\frac{1}{n^2} \sum_{i=1}^{n} \sum_{j=1}^{n} H_{ij} \ln\left(sigmoid(z_i z_j^T)\right) + (1 - H_{ij})(1 - H_{ij} \ln\left(sigmoid(z_i z_j^T)\right) \tag{15}$$

Please note that $z_i$ is the row $i$ vector of the embedding matrix $Z$ and $z_i \in R^{1*D}$. We train the two parameter matrices $\theta_1 \in R^{L_1*L_2}$ and $\theta_2 \in R^{L_2*D}$ using the gradient descent method. Then, partition the samples $(z_i)_{i=1,...,n}$ in $R^{1*D}$ with the k-means algorithm into $k$ clusters/groups. In general, the hypergraph CNN-based clustering technique can be shown in Figure 2.

### 3.4. Discussion about hypergraph convolutional neural network-based clustering technique

From the section 3.3, unlike the graph CNN-based clustering techniques, instead of employing the pairwise relationships among objects/entities/samples, we easily see that this hypergraph CNN-based clustering technique employs the high order relationship among objects/entities/samples. This leads to no information loss. Hence, this hypergraph CNN-based clustering technique is expected to perform better than that of the graph CNN-based clustering technique. The claim is backed in section 4.

However, like the graph CNN-based clustering technique, this hypergraph CNN-based clustering technique is the offline clustering technique. In other words, when new samples arrive, we have to update the incidence matrix of the hypergraph, and we have to re-train our hypergraph auto-encoder.

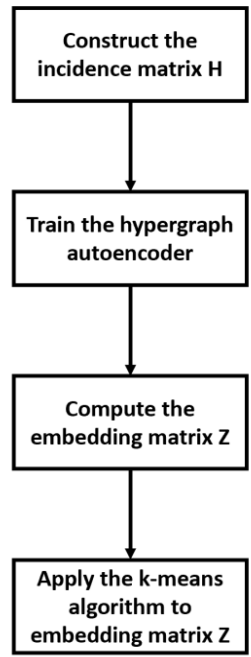

Figure 2. The hypergraph convolutional neural network-based clustering technique

## 4. EXPERIMENTS AND RESULT
### 4.1. Preliminaries on dataset

There are two publicly available dataset, the Citeseer dataset and the Cora dataset. We use them to test our novel clustering technique. For example, we test the graph CNN-based clustering technique.

Cora: This dataset contains 2,708 scientific papers (i.e., nodes of the graph) linked by 5,429 edges representing one scientific paper citing to another. Individually, every paper (i.e., node of the graph) has a binary vector that describes the absence or presence of 1,433 unique words. This binary vector is called the feature vector of the scientific paper. The goal of the experiment is to classify each scientific paper into 7 categories which are Reinforce_Learning, Theory, Rule_Learning, Case_Based, Genetic_Algorithms, Probabilistic_Methods, Neural_Networks, and Theory.

Citeseer: The dataset contains 3,312 scientific papers (i.e., nodes of the graph) linked by 4,732 edges representing one scientific paper citing to another. Individually, every paper (i.e., node of the graph) has a binary vector that describes the absence or presence of 3,703 unique words. This binary vector is called the feature vector of the scientific paper. The goal of the experiment is to classify each scientific paper into one of 6 categories: Human computer interaction (HCI), machine learning (ML), database (DB), information retrieval (IR), artificial intelligence (AI), ML and Agents.

Please note that in the case that the adjacency matrices are not given, the similarity graph need to be built from the feature vectors of the datasets in the following ways: i) The fully connected graph: Connect all samples; ii) The $\varepsilon$-neighborhood graph: All the samples whose pairwise distances are smaller than $\varepsilon$ are connected; and iii) KNN graph: sample $i$ connects with sample $j$ by an edge in a no direction: un-directed graph if sample $i$ is among the $k$ nearest neighbors of sample $j$ or sample $j$ is among the KNN of sample $i$. The KNN graph is used to build the similarity graph from feature vectors of the Cora dataset and the Citeseer dataset.

For notation, we have $k$ set as 5. Finally, the way showing how to construct the hypergraph (i.e., the incidence matrix $H$) from the feature vectors is discussed in detail in section 3. Please note that $D$, the dimensions of the embedding matrix $Z$, is set to be 16.

### 4.2. Experiment results

We compare the performance of the hypergraph CNN-based clustering technique with the performances of the graph CNN-based clustering technique, the k-means clustering technique, the spectral clustering technique for feature datasets, the spectral clustering technique for network datasets. Our model is tested (Python code) with NVIDIA Tesla K80 GPU (12 GB RAM option) on Google Colab. There are three performances of clustering techniques that we are going to employ in this paper which are: i) Silhouette coefficient, ii) Davies-Bouldin score, and iii) Calinski-Harabasz score.

Each sample defines its Silhouette coefficient. The performance consists of two score: i) *a*: In the same groups/clusters, the mean distance between the sample and all other points/samples; and ii) *b*: In the next nearest cluster, the mean distance between the sample and all other points/samples. Then for a single sample, the Silhouette coefficient *s* could be computed in (16):

$$s = \frac{b-a}{\max(a,b)} \tag{16}$$

The Silhouette coefficient for all samples is the silhouette coefficient of the dataset. From (16), we easily recognize that the higher the silhouette coefficient, the better the clustering results. The average similarity measure of each cluster with its most similar cluster, where the ratio of within-cluster distances to between-cluster distances is the similarity, is called The Davies-Bouldin score. For that reason, clusters that are less dispersed and farther apart generate a better score. For notation, zero is the minimum score.

From (16), we easily see that the lower the Davies-Bouldin score, the better the clustering results. The variance ratio criterion or also called the Calinski-Harabasz score is defined as the ratio of the sum of inter-cluster dispersion for all clusters and between-clusters dispersion. From the definition, we easily see that the higher the Calinski-Harabasz score, the better the clustering results. For the Citeseer dataset, Table 1 presents the performances of the graph CNN-based clustering technique, the k-means clustering technique, the spectral clustering technique for feature vectors, the spectral clustering technique for adjacency matrix.

Table 1. Comparison of graph CNN-based clustering technique with other techniques with the Citeseer dataset

| | Silhouette coefficient | Davies-Bouldin score | Calinski-Harabasz score |
|---|---|---|---|
| The hypergraph convolutional neural network-based clustering technique | 0.5034 | 0.4786 | 40369.4615 |
| The graph convolutional neural network-based clustering technique | 0.1203 | 2.0923 | 231.2733 |
| The k-means clustering technique | 0.0007 | 7.3090 | 15.2688 |
| The spectral clustering technique for feature vectors | 0.0047 | 5.9004 | 4.1290 |
| The spectral clustering technique for adjacency matrix | -0.0081 | 14.8230 | 2.1392 |

For the Cora dataset, Table 2 presents the performances many techniques. There is the graph CNN-based clustering technique, the k-means clustering technique, the spectral clustering technique for feature vectors. There is also the spectral clustering technique for adjacency matrix.

Table 2. Comparison of graph CNN-based clustering technique with other techniques with the Cora dataset

| | Silhouette coefficient | Davies-Bouldin score | Calinski-Harabasz score |
|---|---|---|---|
| The hypergraph convolutional neural network-based clustering technique | 0.3349 | 1.0471 | 1288.7860 |
| The graph convolutional neural network-based clustering technique | 0.1963 | 1.7312 | 322.3034 |
| The k-means clustering technique | 0.0149 | 5.5878 | 21.5432 |
| The spectral clustering technique for feature vectors | -0.0196 | 6.6584 | 20.4136 |
| The spectral clustering technique for adjacency matrix | -0.0465 | 8.9685 | 2.0282 |

**4.3 Discussions**

With the results from the Tables 1 and 2, we acknowledge that the graph CNN based clustering technique is superior than the k-means clustering technique, the spectral clustering technique for feature vectors, the spectral clustering technique for adjacency matrix since graph CNN based clustering technique utilize both the information from the feature vectors and the adjacency matrix of the dataset and noises and redundant features in the dataset (i.e., both in the feature vectors and the adjacency matrix) are removed. Moreover, the hypergraph CNN-based clustering technique is better than the CNN-based clustering technique since the hypergraph data structure employs the high order relationships among the samples/entities/objects. This leads to no loss of information.

However, when new samples arrive, for the (hyper)-graph CNN-based clustering techniques, we cannot predict which clusters these samples belong to. That is to say, we have to update the adjacency matrix or the incidence matrix, re-train our (hyper)-graph auto-encoder, re-compute the embedding matrix, and apply k-means clustering technique to this embedding matrix again to get the final clustering result. Hence the (hyper)-graph CNN-based clustering technique can only be considered as the offline clustering technique.

Last but not least, please note that when new samples/data points arrive, we need to re-train our model. For industrial projects, this novel model can be updated/re-trained twice per month or once per month. Let's consider the case when a lot of samples arrive, this event introduces a lot of novel patterns in the whole dataset. Model updating is a must. In the other words, we have to update all the models such as the hypergraph CNN-based clustering technique, the graph CNN-based clustering technique, the k-means clustering technique, the spectral clustering technique for feature vectors, and the spectral clustering technique for adjacency matrix.

## 5. CONCLUSION

The main contributions of our paper are to develop the novel graph CNN-based clustering technique and the novel hypergraph CNN-based clustering technique. We then apply these novel clustering techniques to two Citeseer and Cora datasets and compare the performance of the hypergraph CNN-based clustering technique with that of the graph CNN-based clustering technique, the k-means clustering technique, the spectral clustering technique for feature vectors, the spectral clustering technique for adjacency matrix. In this paper, we presented the (hyper)-graph CNN-based clustering technique to handle the clustering problem. The work is presumably, not complete. Many types of (hyper)-graph CNN are available. In future research, the (hyper)-graph CNN with/without attention could also be used to solve this clustering problem.

## ACKNOWLEDGEMENTS

We would like to thank Ho Chi Minh City University of Technology (HCMUT), VNU-HCM for the support of time and facilities for this study.

## BIOGRAPHIES OF AUTHORS

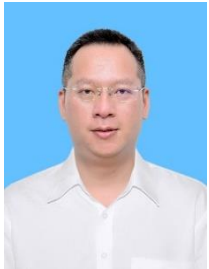 **Linh H. Tran** received the B.S. degree in Electrical and Computer Engineering from University of Illinois, Urbana-Champaign (2005), M.S. and PhD. in Computer Engineering from Portland State University (2006, 2015). Currently, he is working as lecturer at Faculty of Electrical-Electronics Engineering, Ho Chi Minh City University of Technology-VNU HCM. His research interests include quantum/reversible logic synthesis, computer architecture, hardware-software co-design, efficient algorithms and hardware design targeting FPGAs and deep learning. He can be contacted at email: linhtran@hcmut.edu.vn.

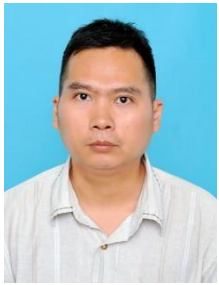 **Loc H. Tran** completed his Bachelor of Science and Master of Science in Computer Science at University of Minnesota in 2003 and 2012 respectively. Currently, he's a researcher at John von Neumann Institute, Vietnam. His research interests include spectral hypergraph theory, deep learning. He can be contacted at email: tran0398@umn.edu.

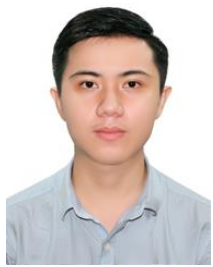 **Nguyen Trinh** received the B.S. and M.S. degree in Electronics Engineering from Ho Chi Minh City University of Technology (HCMUT), Vietnam (2019, 2021). He is also working as lecturer at Faculty of Electrical-Electronics Engineering, Ho Chi Minh City University of Technology-VNU HCM. He can be contacted at email: nguyentvd@hcmut.edu.vn